\documentclass{article}

\usepackage[english]{babel}

\usepackage[letterpaper,top=2cm,bottom=2cm,left=3cm,right=3cm,marginparwidth=1.75cm]{geometry}

\usepackage{amsmath}
\usepackage{graphicx}
\usepackage[colorlinks=true, allcolors=blue]{hyperref}
\usepackage{authblk} 

\title{An Automated Retrieval-Augmented Generation LLaMA-4 109B-based System for Evaluating Radiotherapy Treatment Plans}

\author[1]{Junjie Cui}
\author[1]{Peilong Wang, PhD}
\author[1]{Jason Holmes, PhD}
\author[1]{Leshan Sun, PhD}
\author[2]{Michael L. Hinni, MD}
\author[3]{Barbara A. Pockaj, MD}
\author[1]{Sujay A. Vora, MD}
\author[1]{Terence T. Sio, MD, MS}
\author[1]{William W. Wong, MD}
\author[1]{Nathan Y. Yu, MD}
\author[1]{Steven E. Schild, MD}
\author[1]{Joshua R. Niska, MD}
\author[1]{Sameer R. Keole, MD}
\author[1]{Jean-Claude M. Rwigema, MD}
\author[1]{Samir H. Patel, MD}
\author[1]{Lisa A. McGee, MD}
\author[1]{Carlos A. Vargas, MD}
\author[1]{Wei Liu, PhD\thanks{Corresponding author: \texttt{liu.wei@mayo.edu}}}

\affil[1]{Department of Radiation Oncology, Mayo Clinic Arizona, Phoenix, AZ 85054}
\affil[2]{Department of Otolaryngology, Mayo Clinic Arizona, Phoenix, AZ 85054}
\affil[3]{Department of General Surgery, Mayo Clinic Arizona, Phoenix, AZ 85054}

\begin{document}
\maketitle

\begin{abstract}
\textbf{Purpose}

To develop a retrieval-augmented generation (RAG) system powered by LLaMA-4 109B for automated, protocol-aware, and interpretable evaluation of radiotherapy treatment plans.

\textbf{Methods and Materials}

We curated a multi-protocol dataset of 614 radiotherapy plans across four disease sites and constructed a knowledge base containing normalized dose metrics and protocol-defined constraints. The RAG system integrates three core modules: a retrieval engine optimized across five SentenceTransformer backbones, a percentile prediction component based on cohort similarity, and a clinical constraint checker. These tools are directed by a large language model (LLM) using a multi-step prompt-driven reasoning pipeline to produce concise, grounded evaluations.

\textbf{Results}

Retrieval hyperparameters were optimized using Gaussian Process on a scalarized loss function combining root mean squared error (RMSE), mean absolute error (MAE), and clinically motivated accuracy thresholds. The best configuration, based on \texttt{all-MiniLM-L6-v2}, achieved perfect nearest-neighbor accuracy within a 5-percentile-point margin and a sub-2pt MAE. When tested end-to-end, the RAG system achieved 100\% agreement with the computed values by standalone retrieval and constraint-checking modules on both percentile estimates and constraint identification, confirming reliable execution of all retrieval, prediction and checking steps.

\textbf{Conclusion}

Our findings highlight the feasibility of combining structured population-based scoring with modular tool-augmented reasoning for transparent, scalable plan evaluation in radiation therapy. The system offers traceable outputs, minimizes hallucination, and demonstrates robustness across protocols. Future directions include clinician-led validation, and improved domain-adapted retrieval models to enhance real-world integration.

\end{abstract}

\section{Introduction}

Radiotherapy treatment planning aims to generate a treatment plan that delivers a therapeutic dose to the tumor while minimizing radiation exposure to surrounding healthy tissues \cite{gardner2019modern, deng2021critical, zhang2011parameterization, schild2014proton, li2012dynamically, shan2020intensity}. An essential step in this workflow is the plan evaluation, in which the quality of the proposed dose distribution is assessed. Plan quality refers to the clinical suitability of the dose distribution that a treatment plan can reasonably achieve with optimal coverage of the target by the prescribed dose while minimizing dose to normal tissues \cite{hernandez2020plan}. This is typically evaluated through a quantitative analysis of the dose distribution produced by the treatment planning system (TPS) \cite{geurts2022aapm} or by manual peer review \cite{leone2024visualization}. However, this approach is time-consuming and often involves subjective judgments that may vary between clinicians \cite{leone2024visualization,wang2025feasibility}.

Given the limitations of manual evaluation, extensive efforts \cite{hernandez2020plan, hussein2018automation, cao2022knowledge, hansen2022plan, patel2021radiotherapy, liu2017exploratory, liu2016robustness, kang2019impact, liu2016robustness, liu2023artificial} have been made to develop objective and efficient approaches for radiotherapy plan evaluation using statistical and mathematical frameworks. A common strategy is population-based scoring, in which plan quality is evaluated relative to a cohort distribution. For example, Leone et al.\cite{leone2024visualization} proposed a geometric mean–based scoring system that normalizes each dose–volume histogram (DVH) indices by its clinical limit and ranks treatment plans by their percentile within a protocol-specific cohort. This method demonstrated strong correlation with physician Likert ratings and provided a cohort-aware framework for quality assessment. Other approaches have focused on statistical modeling of DVH distributions. Wahl et al.\cite{wahl2020analytical} introduced a probabilistic model that analytically estimates expected DVHs and $\alpha$-DVHs under delivery uncertainties, enabling robust plan evaluation without the need for Monte Carlo simulation. In a complementary direction, population-derived visual dashboards have been developed to benchmark DVH indices against historical distributions, offering intuitive and interactive feedback via statistical overlays~\cite{mayo2017incorporating}. Meanwhile, mathematical optimization methods have also been employed to assess and improve plan quality directly. Engberg et al.\cite{engberg2017explicit} formulated a convex optimization framework using mean-tail-dose as a risk measure, allowing more precise control over DVH tail behavior. Zhang et al.\cite{zhang2020direct} extended this line of work by treating DVH indices as differentiable functionals, enabling direct optimization of clinically relevant constraints. To facilitate the large-scale application of such frameworks, Pyakuryal et al.~\cite{pyakuryal2010computational} developed the Histogram Analysis in Radiation Therapy (HART) tool, which automates the extraction and statistical analysis of DVH indices, significantly enhancing the reproducibility and efficiency of plan evaluation. Collectively, these works demonstrate that statistical and mathematical methods can provide rigorous, interpretable, and clinically grounded strategies for automated plan evaluation. However, such approaches often require hand-engineered metrics, are limited to predefined protocols, and may not generalize well across institutions with different naming conventions or contouring practices \cite{hernandez2020plan, hansen2022plan}. Furthermore, their outputs tend to be static and lack the interactive, explanatory capabilities needed to support dynamic clinical workflows or accommodate evolving clinical guidelines. These limitations have motivated interests in more flexible, learning-based systems, particularly those powered by large language models (LLMs), which can reason over heterogeneous inputs, provide interpretable summaries, and integrate auxiliary tools to support complex decision-making \cite{mesinovic2025explainability, hu2025systematic}.

\begin{figure}[t]
    \centering
    \includegraphics[width=\textwidth]{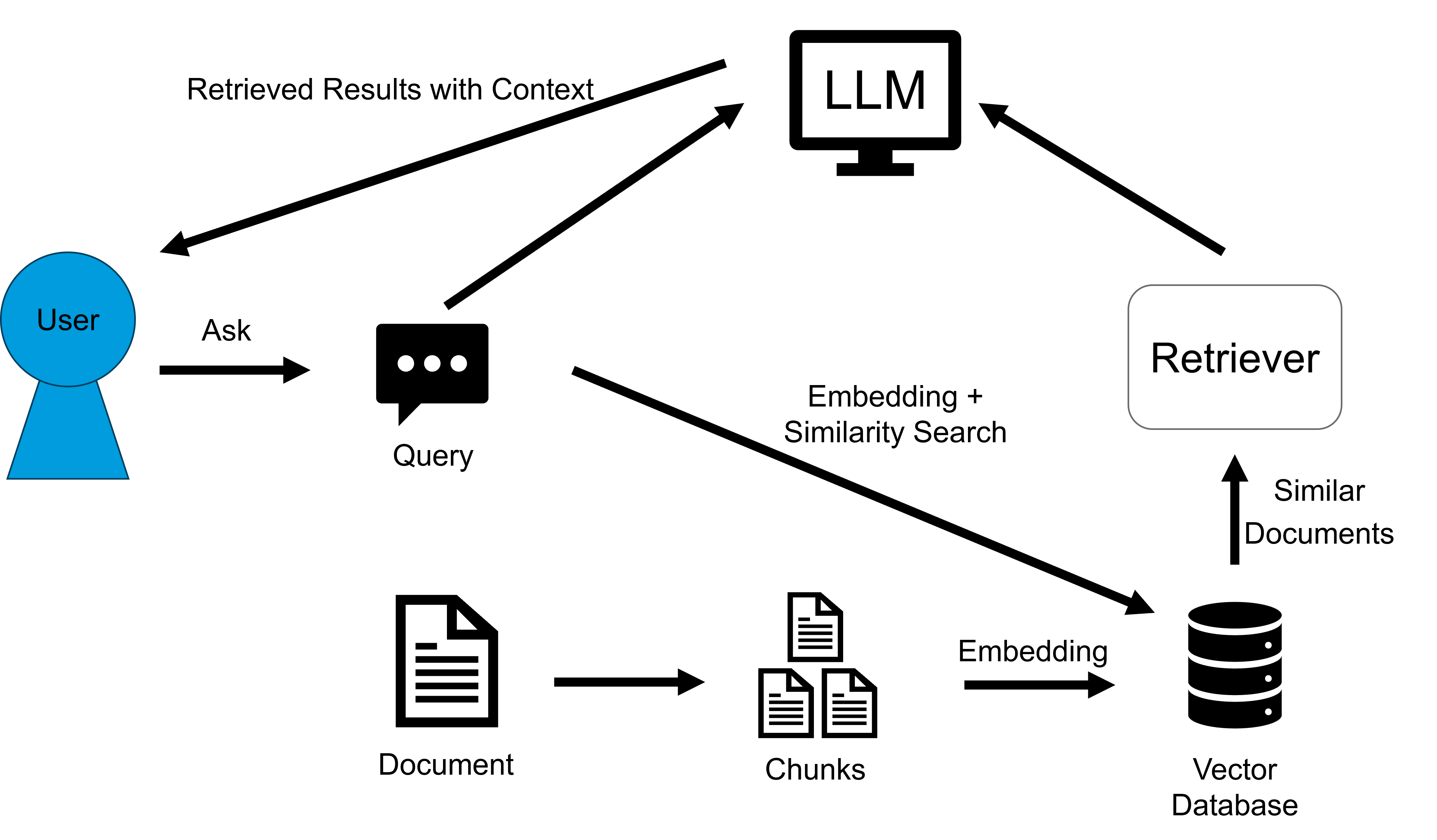}
    \caption{Overview of the retrieval-augmented generation (RAG) workflow. A user query is processed by a retriever to fetch relevant context from a knowledge base, which is then fed into an LLM for grounded generation.}
    \label{fig:rag}
\end{figure}

Recent advances in LLMs, a class of deep learning models trained on massive corpora of texts, have opened new opportunities for automating complex clinical workflows. LLMs are capable of performing natural language understanding, multi-turn reasoning, and code execution, and can interface with external tools to support decision-making tasks \cite{wang2023mint, RN1787, RN1782, RN1790, RN1753, RN1169, RN1789, RN1559, RN1553, RN1389, RN1785}. These capabilities arise from their ability to learn high-dimensional representations of language and generalize across diverse inputs without task-specific supervision \cite{budnikov2025generalization}. As a result, LLMs are increasingly being explored for medical applications that require interpreting clinical text, synthesizing structured and unstructured data, and generating human-readable summaries \cite{goodell2025large,nusrat2025autonomous}. In particular, retrieval-augmented generation (RAG) systems combine the generative capabilities of LLMs with external knowledge retrieval mechanisms, allowing the model to incorporate relevant, case-specific contexts into its reasoning process \cite{amugongo2025retrieval}. A typical RAG workflow is illustrated in Fig.~\ref{fig:rag}, where the user query triggers a retrieval step that fetches relevant documents, which are then passed to the LLM for grounded response generation. This paradigm has shown promise in domains like clinical decision support and radiology report generation, and is now beginning to impact radiotherapy planning \cite{wada2025retrieval}. For instance, a recent feasibility study by Wang et al.\cite{wang2025feasibility} demonstrated that GPT-4-based agents could partially automate the radiotherapy treatment planning process, including dose prescription, structure prioritization, and plan evaluation, using chain-of-thought prompting and self-verification. Similarly, Liu et al.\cite{liu2024automated} explored a multi-agent LLM system for guiding the full radiotherapy pipeline, including image interpretation and treatment planning, through GPT-4 Vision and command-executing submodules. Although these systems aim to automate the entire planning process, their results highlight the importance of accurate and interpretable plan evaluation as a critical subtask. In parallel, domain-specific applications of RAG systems have emerged, in which LLMs are guided by retrieved dose-volume metrics and clinical protocols to assess new treatment plans against historical cohorts \cite{nusrat2025autonomous,xu2024crp}. These approaches represent a promising direction for building clinically informed, interpretable, and adaptable tools for radiotherapy plan evaluation.

To address the need for accurate, interpretable and scalable plan evaluation, we introduce an automated RAG system powered by LLaMA-4 109B for radiotherapy quality assessment. Our framework integrates (1) a scoring module that computes normalized dose metrics and population-based percentiles, (2) a retrieval module that identifies similar historical plans based on numerical and textual features, and (3) a constraint-checking tool that flags clinical violations using protocol-defined thresholds. These components are leveraged through an LLM-driven reasoning process that issues explicit tool calls to retrieve context, perform checks, and synthesize a structured, protocol-aware summary of plan quality. Unlike prior end-to-end systems, our method separates data, logic, and generation, reducing hallucinations while supporting traceability and flexible protocol integration. The output includes a quantitative plan score and a list of failed constraints, enabling transparent decision support aligned with clinical practice. We curate a multi-protocol knowledge base from 614 historical plans and demonstrate the system’s accuracy and reliability through evaluation.

\section{Method}
    All research activities of this study were conducted under institutional review board (IRB) approval. The IRB number associated with this study is 24-010322: ``Application of Large Language Models (LLMs) to Enhance Efficiencies of Clinical and Research Tasks in Radiation Oncology".
    
    \subsection{Dataset and Clinical Protocols}

        We curated a dataset of 614 radiotherapy treatment plans across 463 patients, covering four disease sites: head and neck (3 protocols, 175 plans), prostate (3 protocols, 264 plans), breast (2 protocols, 136 plans) and lung (1 protocol, 48 plans). All cases were drawn from clinical trials conducted at the Mayo Clinic Arizona. We focused on clinical trial data because such plans are consistently annotated and strictly follow protocol-defined dose constraints for target regions and organs-at-risk (OARs), making them ideal for training and evaluating a protocol-aware system.

        Initial identification of eligible plans was facilitated using RadOnc-GPT~\cite{liu2023radonc}, a domain-specialized large language model capable of retrieving unstructured oncology information from electronic records and assisting in clinical cohort discovery. Each plan included a protocol-defined set of mandatory DVH endpoints, which were used to guide evaluation. A full breakdown of the protocols and the corresponding DVH endpoints is provided in Table~\ref{tab:protocol_dvh}.

        \begin{table}[ht]
        \centering
        \footnotesize
        \begin{tabular}{|p{3 cm}|p{7cm}|}
        \hline
        \textbf{Protocol Name} & \textbf{Required DVH Endpoints} \\
        \hline
        Lung 1 & Esophagus D33\%, D67\%, D100\%; Liver D100\%, D50\%; Cord Max; Heart D33\%, D67\%, D100\%, Mean, V50Gy; Skin Max; Lung Total V20Gy \\
        \hline
        Head \& Neck 1 & Cord Max; Brain Stem Max; Lips Mean; Oral Cavity Mean; Parotid Mean; Esophagus Mean; Submandibular Mean; Larynx Mean \\
        \hline
        Head \& Neck 2 & Brain Stem Max; Parotid/Submandibular/Pharyngeal/ Larynx/Oral Cavity Mean; Mandible Max; Brachial Plexus Max \\
        \hline
        Head \& Neck 3 & Cord D0.03cc; Brain Stem D0.03cc; Lips/Oral Cavity/Parotid/Submandibular/Larynx Mean; OAR Pharynx Mean, V15\%, V33\%; Cervical Esophagus Mean \\
        \hline
        Breast 1 & Heart D0.01cc, Mean; Breast Skin D0.01cc; Lung Ipsilateral/Contralateral V40\%; Esophagus D0.01cc; Brachial Plexus D0.01cc \\
        \hline
        Prostate 1 & Rectum V65\%, V90\%, D0.03cc; Bladder V90\%, D0.03cc; Femoral Heads (L/R) V50cc \\
        \hline
        Prostate 2 & Rectum V30Gy; Bladder V30/33cc; Femoral Heads V40cc; Small/Large Bowel Max; Lung Total V30Gy \\
        \hline
        Prostate 3 & (Multiple arms) Rectum/Bladder V40--70Gy, Dmax; Femoral Heads V40cc or Max \\
        \hline
        Breast 2 & (Photon/Proton arms) Heart Max, Mean; Breast Skin Max; Lung Ipsilateral/Contralateral V50\% \\
        \hline
        \end{tabular}
        \caption{Summary of clinical protocols used in this study, including disease sites, number of patients, number of treatment plans, and the protocol-defined DVH indices used for evaluation.}
        \label{tab:protocol_dvh}
        \end{table}
        
        Patient demographics for the full cohort are summarized as follows: the mean age was 64.8 years (range: 29–91; standard deviation (SD): 10.4). The majority of patients identified as White (91.4\%), with the remainder distributed across Black or African American (2.2\%), Native American (2.6\%), Asian (2.6\%), and other or undisclosed categories (1.3\%).

        To preprocess the data, we first extracted the protocol-defined dose volume constraints directly from official protocol documents housed in the institution’s clinical trial database. Plans were organized and saved into separate text files based on their protocol assignments, ensuring that each file contained only the relevant dose metrics and identifiers for downstream evaluation. We then parsed each treatment plan to extract the DVH indices required by its associated protocol. This structured preprocessing step enabled streamlined normalization and scoring in subsequent stages of system development.
        
    \subsection{Knowledge Base–Enabled Plan Scoring}
        Our plan evaluation framework is inspired by previous work on population-based scoring by Leone et al.~\cite{leone2024visualization}, which showed that aggregating protocol-defined DVH indices can yield clinically interpretable assessments of plan quality. In our system, each clinical protocol specifies a set of DVH indices that must be evaluated for specific OARs or targets. These indices may include metrics such as D33\% (the minimum dose received by the hottest 33\% of a structure’s volume) and V50Gy (the percentage of a structure’s volume receiving at least 50 Gray). These metrics reflect both high-dose regions and volumetric coverage, and are widely used to ensure both tumor control and healthy tissue sparing. 
        
        For each treatment plan, we extract the raw values of all protocol-required DVH indices. To allow fair aggregation across metrics with different units and scales, each raw value is then normalized by dividing it by its corresponding clinical constraint (e.g., a dose threshold or volume limit), and multiplied by 100 to express it as a percentage:
        \[
        \text{normalized score}_i = \frac{\text{raw value}_i}{\text{constraint limit}_i} \times 100 + \varepsilon.
        \]
        Here, $\varepsilon = 10^{-6}$ is a small constant added to prevent zero values and numerical instability. 
        
        To obtain a single summary score for the entire plan, we compute the geometric mean of the normalized values across all DVH indices:
        \[
        \text{gm\_score} = \left( \prod_{i=1}^{n} \text{normalized score}_i \right)^{1/n}.
        \]
        This approach ensures that no single large violation is masked by low values in other metrics, as the geometric mean is more sensitive to outliers than an arithmetic mean. It also naturally balances all constraints without requiring hand-tuned weights. 
        
\begin{figure}[ht]
            \centering
            \includegraphics[width=\textwidth]{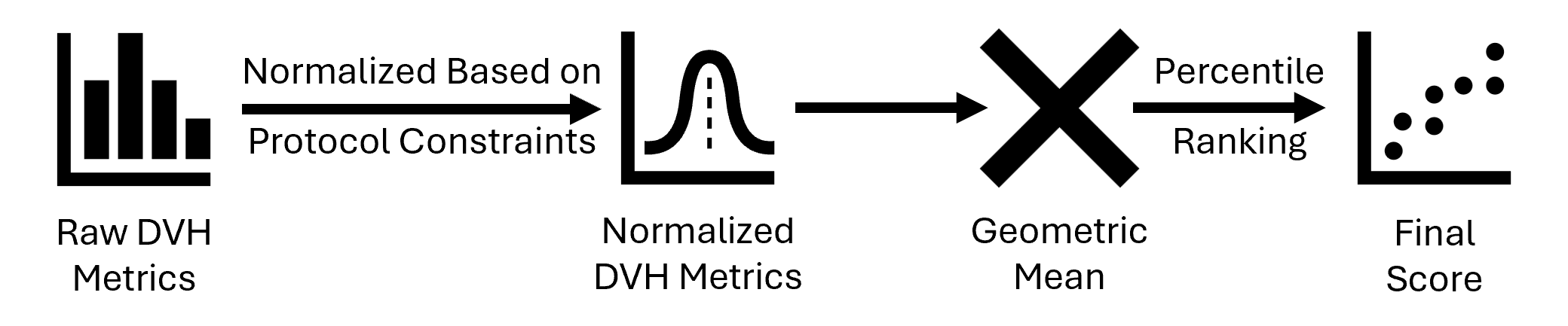}
            \caption{Overview of the plan scoring pipeline. For each treatment plan, protocol-defined DVH indices (e.g., D33\%, V50Gy) are extracted and normalized against their corresponding protocol constraints. These normalized values are aggregated using the geometric mean to compute a single value, which is then mapped to a percentile rank within the protocol-matched cohort. The percentile rank is the final score.}
            \label{fig:scoring_pipeline}
        \end{figure}

        Finally, the geometric mean score is mapped to a percentile rank within the historical cohort of plans for the same protocol. This percentile serves as the final plan score and enables direct comparison across different plans operating under shared clinical standards. A schematic overview of the full plan scoring pipeline, from raw DVH indices to percentile-based plan scores, is illustrated in Fig.~\ref{fig:scoring_pipeline}.

        To support retrieval and reasoning, each scored plan is stored as a JavaScript Object Notation (JSON) entry containing its protocol name, raw DVH indices, normalized metrics, geometric mean, and percentile rank. To enable evaluation of retrieval and generalization, the plans were partitioned by protocol: 90\% were used to build the knowledge base (579 plans) for our RAG system, while the remaining 10\% were held out as a testing dataset (62 plans). For testing plans, percentile scores were hidden at the inference time, allowing the RAG system to generate plan-level assessments using retrieved examples and protocol-based reasoning.
        
    \subsection{Retrieval and Prediction}
        To predict the clinical quality of a new radiotherapy treatment plan, we designed a retrieval-based system that identifies similar historical plans from the knowledge base and uses their percentile scores to estimate the quality of the input plan. For each protocol, we constructed three similarity indexes over the historical plans: (1) a text-based index encoding a natural language representation of each plan’s DVH indices using sentence embeddings; (2) a normalized-metric index built from numerical vectors of protocol-defined metrics, each normalized as a percentage of its clinical limit; and (3) a raw-metric index constructed using the unnormalized DVH index values.

        At the inference time, given a new test plan, we first retrieve the top $k$ plans with the most similar geometric mean scores from the same protocol. These initial candidates are then re-ranked using a weighted similarity score that combines three components: semantic similarity between the textual descriptions of metric values, vector similarity of normalized metric profiles, and vector similarity of raw metric values. The similarity weights are tunable, allowing flexible control over the influence of each component. For each test plan, we report the nearest neighbor’s percentile score, as well as a weighted average and weighted median of the retrieved neighbors’ percentiles, which together form the final predicted quality estimates. This multi-view retrieval strategy allows the system to generalize across protocols and capture both numerical and semantic patterns in historical plan data.
        
    \subsection{RAG System for Plan Evaluation}
\begin{figure}[ht]
            \centering
            \includegraphics[width=\textwidth]{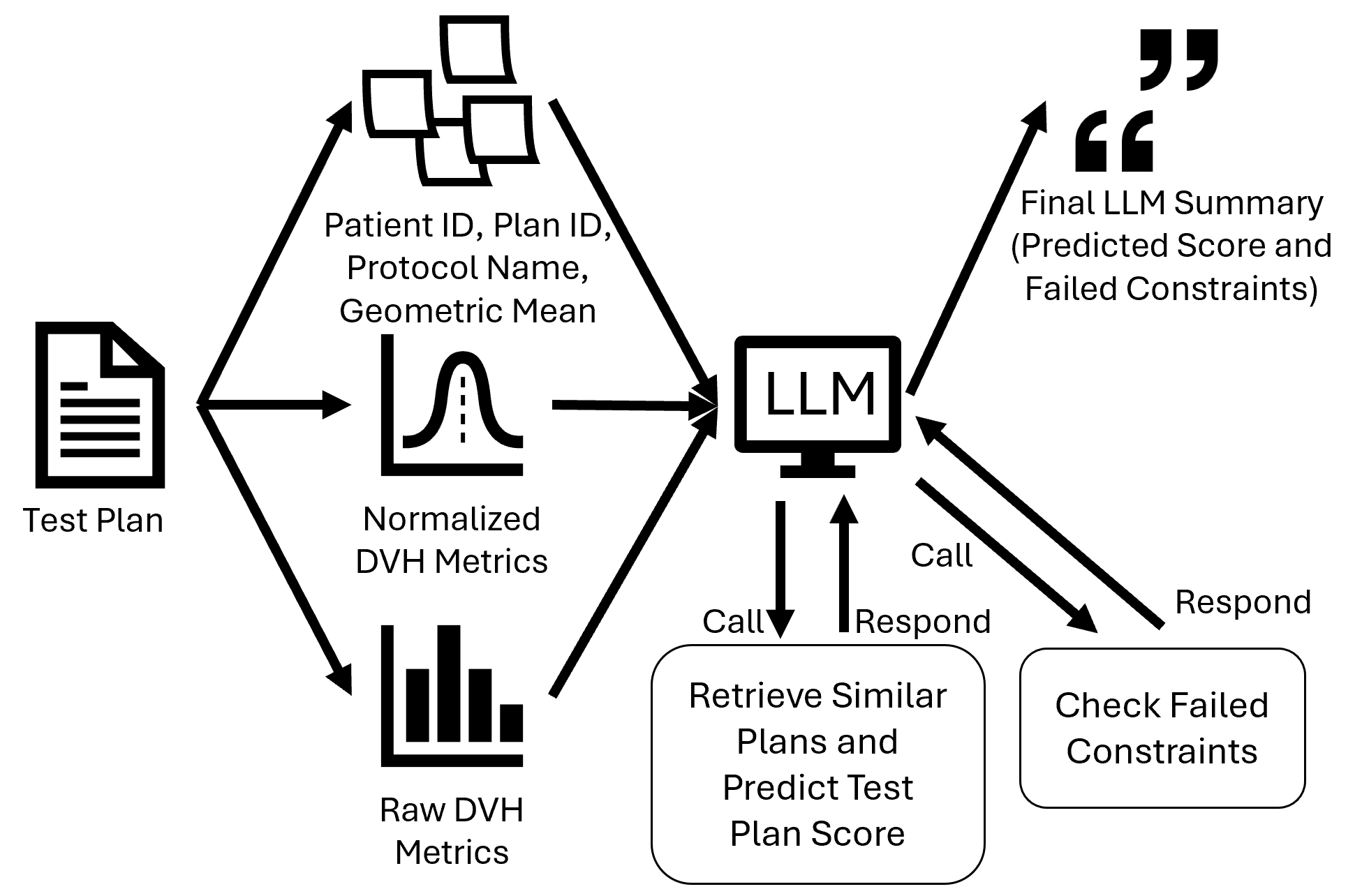}
            \caption{Overview of our RAG system workflow for radiotherapy plan evaluation. Given a new treatment plan, the system first computes DVH indices normalized to the protocol-defined constraints and the geometric mean of the normalized dose metrics. The geometric mean, the normalized and raw DVH indices are used to query a retrieval module, which retrieves similar historical plans from a protocol-matched knowledge base, and makes predictions on the test plan quality based on the retrieved historical plans. In parallel, a constraint-checking module identifies violations based on the protocol constraints. The retrieved plans and constraint results are then passed to LLaMA 4 109B, which synthesizes a concise and interpretable summary describing the percentile-based plan quality and any failed constraints.}
            \label{fig:rag_workflow}
        \end{figure}

        We developed an RAG system powered by LLaMA 4 109B to support the interpretation and evaluation of radiotherapy treatment plans. The system integrates structured dose-volume data with protocol knowledge to produce concise and clinically relevant assessments. An overview of this workflow is illustrated in Fig.~\ref{fig:rag_workflow}.

        Upon receiving a new treatment plan, the system first computes dose metrics normalized to the protocol-defined constraints, along with a summary score reflecting overall plan quality. This structured representation serves as input to the RAG system. The LLM is equipped with two auxiliary tools: a retrieval module that identifies similar historical plans based on both textual and numerical features, and a constraint-checking module that flags violations of protocol constraints.
        
        The system operates in a multi-step, tool-augmented reasoning process. First, the LLM queries the retrieval module to obtain the percentile estimates for the input plan, including its similarity-based nearest neighbor, weighted average and median within the protocol-matched cohort. Next, it invokes the constraint-checking tool to determine which dose-volume metrics exceed clinical constraints. With this contextual information, the LLM then generates a brief and interpretable summary describing the plan’s percentile-based standing and any constraint violations.
        
        This design allows the LLM to produce grounded, protocol-aware evaluations while minimizing hallucinations. By combining structured retrieval, constraint analysis, and generative reasoning in a unified framework, the RAG system enables interpretable and scalable plan evaluation across diverse clinical protocols.

    \subsection{Experimental Design}
        \subsubsection{Experiment 1: Retrieval Accuracy and Ranking Quality}
            The goal of our first experiment was to identify the optimal retrieval strategy for the RAG system by evaluating multiple SentenceTransformer backbones and tuning the associated retrieval related configurations. SentenceTransformer contains a library of pretrained language models used to convert text (e.g., structure names) into dense numerical vectors (embeddings) that capture semantic similarity. These embeddings allow for comparisons between textual features based on cosine similarity or other distance metrics in the latent space. The backbone architecture affects how well these embeddings capture clinical semantics, making it a critical design choice for retrieval tasks involving structured radiotherapy data.
            
            Each SentenceTransformer backbone was evaluated under an independently optimized retrieval configuration. Specifically, we tuned the similarity weighting coefficients for the three retrieval components: the weighting coefficients for textual similarity ($\alpha$), normalized metric distance ($\beta_{\text{norm}}$), and raw metric distance ($\beta_{\text{raw}}$). These weights control the contribution of each feature type when computing similarity between the query and candidate plans. In addition, we tuned the retrieval depth $k$, which specifies the number of nearest neighbors retrieved to form the contextual cohort for percentile estimation. We restricted $k$ to the range of 3 to 10 for methodological and practical reasons. Setting $k = 1$ would bypass any aggregation and rely solely on the nearest neighbor, making it incompatible with our weighted mean and median calculations. A cohort size of $k = 2$ similarly lacks statistical robustness for percentile estimation. On the other end, increasing $k$ beyond 10 risks incorporating weakly similar plans due to the nature of percentile-based scoring and the limited size of our dataset. 

            To optimize these hyperparameters, we employed Gaussian Process (GP) minimization, a sample-efficient black-box optimization method. In our setting, each evaluation of the retrieval configuration requires computing multiple prediction metrics over all test plans, which is computationally expensive. GP optimization is particularly suited to such scenarios: rather than exhaustively testing all combinations of weights and retrieval depths, it builds a probabilistic model that estimates how changes in the hyperparameters affect performance. It then intelligently selects the next configuration to try by balancing the exploration of new regions of the parameter space with the exploitation of known high-performing regions. This approach allowed us to efficiently search for the best combination of similarity weights ($\alpha$, $\beta_{\text{norm}}$, $\beta_{\text{raw}}$) and retrieval depth $k$ for each SentenceTransformer backbone, minimizing the need for brute-force tuning while ensuring that the selected configurations yield clinically accurate percentile predictions.

            The objective for optimization was a scalarized loss function that integrates four clinically motivated evaluation metrics: (1) root mean squared error ($\text{RMSE}_{\text{AVG}}$) between the predicted and true percentiles using weighted average across $k$ retrieved plans; (2) mean absolute error ($\text{MAE}_{\text{NN}}$) between the predicted and true percentiles using the single nearest neighbor; (3) percentage of cases where the 1-nearest-neighbor prediction is within 5 percentile points of the true percentile score (\%$\leq$5pt$_{\text{NN}}$); and (4) percentage of cases where the weighted average prediction is within 10 percentile points (\%$\leq$10pt$_{\text{AVG}}$). The loss function is defined as the sum of RMSE, MAE, and the complement of the two accuracy metrics (normalized to the $[0,1]$ scale): 
            \[\mathcal{L} = \text{RMSE}_{\text{AVG}} + \text{MAE}_{\text{NN}} + \frac{100 - \%\leq 5\text{pt}_{\text{NN}}}{100} + \frac{100 - \%\leq 10\text{pt}_{\text{AVG}}}{100}\]
            This penalizes both large errors and clinically unacceptable predictions. This optimization process was repeated independently for each model to ensure a fair comparison under individually tuned settings.

            Notably, this tuning process did not involve any fine-tuning of model weights for the SentenceTransformer or LLM. Instead, our approach treats these components as frozen modules and focuses on tuning the retrieval logic, similarity weights, and the system prompt. This modularity supports generalizability, ease of deployment, and flexible integration into clinical environments.

            For the best-performing retrieval model, we report a suite of evaluation metrics to assess both the accuracy and clinical relevance of percentile predictions. These include mean absolute error (MAE), root mean squared error (RMSE), Pearson correlation coefficient ($r$), Spearman rank correlation coefficient ($\rho$), coefficient of determination ($R^2$), and the percentage of predictions within 5 and 10 percentile points of the true percentile score. MAE captures the average absolute difference between the predicted and actual percentiles, providing an intuitive sense of the typical deviation encountered in practice. RMSE penalizes larger errors more heavily and thus helps highlight occasional outlier mismatches that may have greater clinical impact. Pearson $r$ quantifies the linear correlation between the predicted and true percentiles, offering insights on how well the model preserves the overall trend in plan quality. Meanwhile, Spearman $\rho$ assesses monotonicity and rank consistency, which is especially useful in clinical decision-making scenarios where plans are selected or triaged based on relative performance rather than precise numeric scores. The value of $R^2$ reflects the proportion of variance in the true percentile score explained by the model, providing a high-level measure of its predictive power. Finally, we report the percentage of test cases, where the predicted percentile falls within 5 or 10 points of the true value: two thresholds that reflect common clinical tolerances for stratifying plan quality. These thresholds offer a pragmatic view of the system’s ability to deliver clinically acceptable estimates under real-world scenarios. Collectively, these metrics offer a multidimensional perspective on retrieval performance, encompassing both statistical rigor and clinical interpretability, and support our selection of the best-performing SentenceTransformer backbone model as the default retriever for our RAG-based plan evaluation system.
            
        \subsubsection{Experiment 2: Evaluation Quality of the RAG System}
            The second experiment aimed to evaluate whether the RAG system, when integrated with LLaMA 4 109B, could produce outputs consistent with those obtained from the individual scoring, retrieval and constraint-checking modules. Prior to the full-scale evaluation, we refined the system prompts to ensure that the LLM followed the intended sequence of the tool calls and returned outputs in the expected format. Once the prompts were finalized, we executed the complete pipeline for each test plan, including plan scoring, retrieval using the optimal configuration from Experiment 1, and LLM-based reasoning in a batch processing setting. The generated summaries were then collected for evaluation.

            To assess the system accuracy, we compared the percentile estimates reported by the LLM (nearest neighbor, weighted average, and weighted median) with the reference values computed directly from the underlying retrieval and scoring components. Consistency was determined by parsing the model responses and automatically identifying discrepancies. In parallel, we evaluated the integration of the system with the constraint checking module by verifying that the LLM correctly invoked the relevant tool call and that the returned set of the violated constraints matched those identified.

            This experiment confirms that the RAG system reliably executes the intended sequence of operations (retrieval, prediction, and constraint checking) and generates outputs aligned with expected outputs in real-world clinical scenarios. 
            
\section{Results}
    \subsection{Plan Retrieval Accuracy}

        We evaluated five SentenceTransformer backbone models to identify the optimal backbone model for retrieving clinically similar plans in our RAG system.  Each model differs in architecture, size, and pretraining corpus. Therefore, finding the model that fits our system is crucial. For this study, we selected models spanning a range of tradeoffs between performance and efficiency: \texttt{all-MiniLM-L6-v2} and \texttt{all-mpnet-base-v2} are smaller and faster, making them more suitable for real-time applications; \texttt{stsb-roberta-large} and \texttt{all-distilroberta-v1} are slower but potentially more accurate because they process languages in greater detail; and \texttt{msmarco-distilbert-base-tas-b} is optimized for retrieval tasks. All models were used without fine-tuning.
        
        To ensure fair comparison, we independently optimized each model's retrieval configuration using the method described in Experiment 1. This involved selecting the optimal similarity weights and retrieval depth $k$ using GP based minimization, targeting a combined loss function over four clinically motivated prediction metrics. 
        
       Table~\ref{tab:retrieval_summary} summarizes the best configurations for each model along with their retrieval accuracy. Despite the technical differences among backbone models, all achieved perfect accuracy in retrieving a nearest neighbor within 5 percentile points of the true plan score. However, \texttt{all-MiniLM-L6-v2} achieved the lowest overall scalarized loss (5.53), reflecting slightly better prediction accuracy and robustness. In practice, this model offers a compelling balance between performance and computational cost, requiring only a fraction of the memory and runtime of larger models such as RoBERTa-large. This makes it a strong candidate for integration into clinical systems. We therefore selected \texttt{MiniLM} for our default SentenceTransformer backbone.
        
        \begin{table}[ht]
        \centering
        \footnotesize
        \resizebox{\textwidth}{!}{%
        \begin{tabular}{|l|c|c|c|c|c|c|c|c|c|}
        \hline
        \textbf{Model} & $k$ & $\alpha$ & $\beta_{\text{norm}}$ & $\beta_{\text{raw}}$ & \textbf{RMSE$_{\text{AVG}}$} & \textbf{MAE$_\text{NN}$} & \%$\leq$5pt$_\text{NN}$ & \%$\leq$10pt$_\text{AVG}$ & \textbf{Loss} \\
        \hline
        \texttt{all-mpnet-base-v2} & 4 & 0.000080 & 0.999793 & 0.000127 & 3.777993 & 1.770161 & 100.0 & 98.387097 & 5.564283 \\
        \texttt{all-MiniLM-L6-v2} & 4 & 0.004313 & 0.983081 & 0.012606 & 3.777999 & 1.738065 & 100.0 & 98.387097 & 5.532193 \\
        \texttt{all-distilroberta-v1} & 4 & 0.000045 & 0.999909 & 0.000045 & 3.777993 & 1.770161 & 100.0 & 98.387097 & 5.564284 \\
        \texttt{msmarco-distilbert-base-tas-b} & 4 & 0.000045 & 0.999909 & 0.000045 & 3.777997 & 1.770161 & 100.0 & 98.387097 & 5.564287 \\
        \texttt{stsb-roberta-large} & 4 & 0.000076 & 0.999879 & 0.000045 & 3.778000 & 1.738065 & 100.0 & 98.387097 & 5.532194 \\
        \hline
        \end{tabular}%
        }
        \caption{Optimized retrieval configurations and summary metrics across the SentenceTransformer backbone models. The best-performing model is \texttt{all-MiniLM-L6-v2}.}
        \label{tab:retrieval_summary}
        \end{table}

        To further evaluate the selected retrieval model (\texttt{MiniLM}), we examined its prediction performance using three common aggregation strategies: nearest neighbor, weighted average, and weighted median. As shown in Table~\ref{tab:retrieval_miniLM}, all models achieved high accuracy. The nearest neighbor approach delivered the lowest mean absolute error (MAE = 1.74) and perfect agreement (100\%) within 5 percentile points of the true percentile score, indicating strong consistency for individual predictions. While the weighted average method had slightly higher MAE and RMSE values, it achieved the best ranking consistency, as reflected by the highest Spearman correlation ($\rho$ = 0.9902). This means that the predicted percentiles followed the same order as the true percentiles more faithfully, which is especially important when treatment plans are selected or prioritized based on the relative ranking.

        \begin{table}[ht]
        \centering
        \footnotesize
        \begin{tabular}{|l|c|c|c|c|c|c|c|}
        \hline
        \textbf{Method} & \textbf{Pearson $r$} & \textbf{Spearman $\rho$} & \textbf{MAE} & \textbf{RMSE} & $R^2$ & \%$\leq$5pt & \%$\leq$10pt \\
        \hline
        Nearest Neighbor   & 0.9971 & 0.9959 & 1.7381 & 1.9013 & 0.9942 & 100.0 & 100.0 \\
        Weighted Average   & 0.9895 & 0.9902 & 2.1361 & 3.7780 & 0.9769 & 91.94 & 98.39 \\
        Weighted Median    & 0.9944 & 0.9943 & 1.9986 & 2.7962 & 0.9874 & 90.32 & 100.0 \\
        \hline
        \end{tabular}
        \caption{Prediction performance of the selected retrieval model (\texttt{all-MiniLM-L6-v2}) across multiple aggregation strategies.}
        \label{tab:retrieval_miniLM}
        \end{table}

        Overall, these results confirm that our retrieval-based prediction pipeline can identify high-quality reference plans with remarkable accuracy, often within 1–2 percentile points of the true percentile scores. This demonstrates the effectiveness of semantically enhanced plan retrieval for cohort-aware, protocol-specific evaluation.
        
        \subsection{LLM-Based Plan Evaluation Accuracy}
            
            In the Experiment 2, we evaluated whether the full RAG system integrating retrieval, percentile prediction, and constraint checking could produce outputs consistent with values obtained from the individual modules. For each test plan, the system was run end-to-end in the batch mode.

            We verified that the LLM executed the expected sequence of operations. Specifically, it retrieved relevant plans using the optimized configuration from Experiment 1, generated percentile predictions based on the retrieved values, and subsequently triggered the constraint-checking tool to identify any violations of clinical constraints. These tool calls were implicitly embedded in the LLM's response, and their outcomes were parsed and compared against the values computed by the standalone retrieval and constraint-checking modules.

\begin{figure}[ht]
                \centering
                \includegraphics[width=\textwidth]{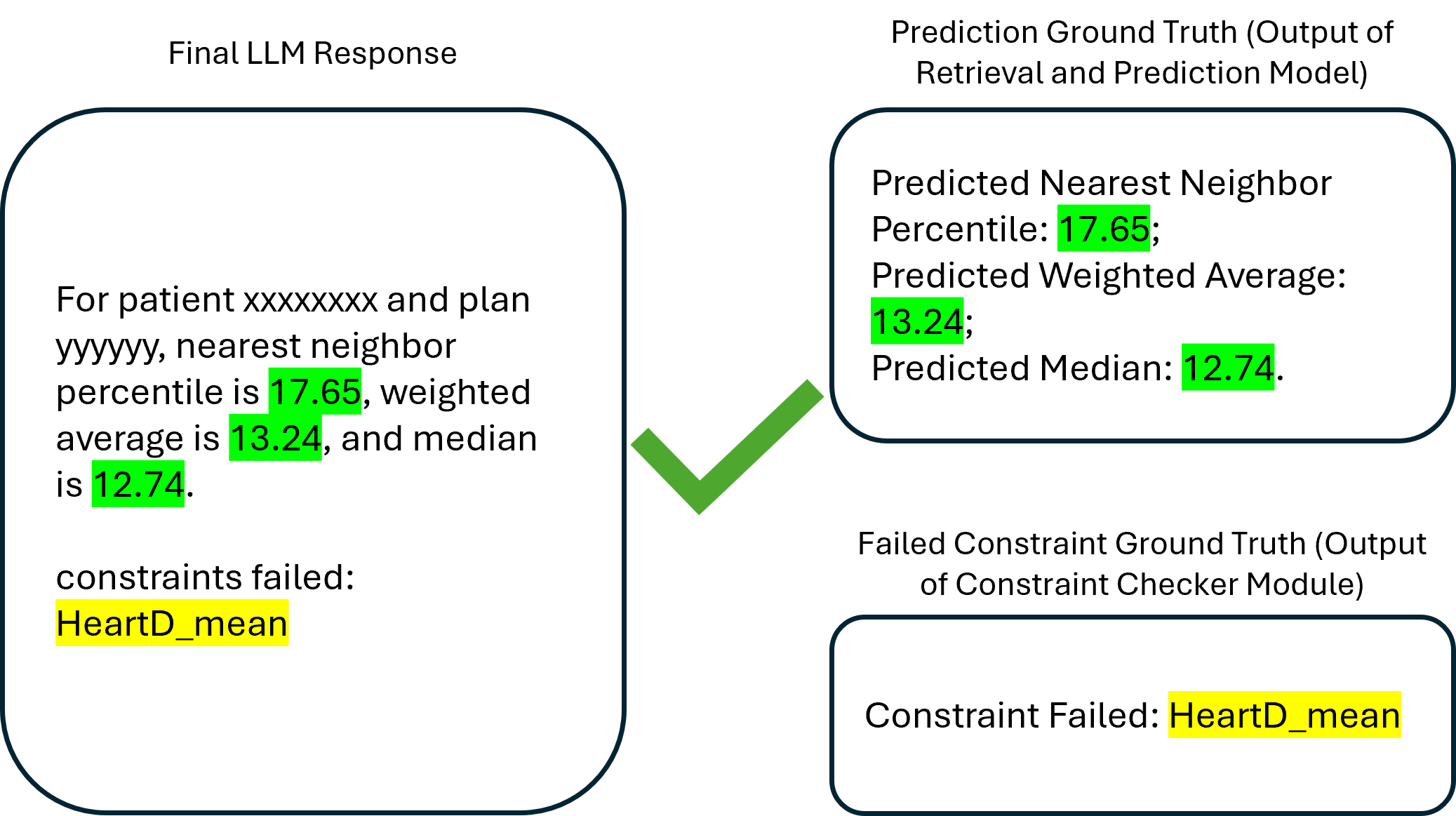}
                \caption{An example demonstrating consistency between the LLM-generated summary (left) and the parsed outputs (right), which include the predicted percentile scores (nearest neighbor, weighted average, and weighted median) and a flagged constraint violation (Heart\_mean dose (Gy)). All outputs match the values produced by the underlying retrieval and constraint-checking modules, confirming faithful execution and integration within the RAG pipeline.}
                \label{fig:llm_response}
            \end{figure}

            For all evaluated plans, the LLM-generated summaries yielded exact matches with the reference values, achieving 100\% agreement with both predicted percentiles (nearest neighbor, weighted average, and weighted median) and clinical constraint violations. This demonstrates that the system not only performs correct retrieval and scoring but also integrates tool outputs into a coherent and clinically interpretable narrative. Fig.~\ref{fig:llm_response} illustrates one such example. On the left, we show the full summary generated by the LLM in response to a test plan. On the right, we display the individual components parsed from the output: the three predicted percentile estimates and the reported constraint violations for the heart structure. All values are in perfect agreement with the computed values from individual modules, highlighting the reliability and transparency of our RAG system in mimicking the behavior of modular pipelines while producing human-readable summaries.

\section{Discussion}
    This work presents an LLaMA 4 109B-based RAG system for automating radiotherapy treatment plan evaluation. We found that our system accurately retrieves clinically similar plans, predicts percentile values with high fidelity, and identifies violated constraints in full agreement with expected outputs. Notably, the LLM demonstrated perfect consistency with independently verified retrieval and constraint-checking tools, achieving 100\% accuracy across all test cases.
    
    \subsection{Key Observations from Retrieval Tuning}
    
        The results of Experiment 1 provide key insights into how percentile prediction accuracy can be optimized for radiotherapy plan evaluation. Although all five SentenceTransformer backbone models performed similarly on standard error-based metrics (e.g., RMSE, MAE), our optimized retrieval configurations consistently prioritized structured and numerical DVH indices over textual metadata. Specifically, the weight assigned to textual similarity between structure names was negligible across models, while the majority of the retrieval signal came from the normalized differences in dose metrics. This suggests that in our dataset, plan identifiers and structure labels did not offer sufficient variation or semantic richness to meaningfully guide retrieval. In contrast, numerical features directly reflected inter-patient variation in treatment delivery and were more effective for identifying clinically similar plans. An important takeaway is the strength of the nearest neighbor method for generating percentile estimates. This simple approach not only achieved the lowest prediction error, but also demonstrated perfect agreement within $\pm 5$ and $\pm 10$ percentile thresholds. These results suggest that, in contexts where a library of similar plans is available, retrieving the most relevant prior case can yield clinically acceptable and highly interpretable benchmarks without requiring complex aggregation or averaging. Together, these findings reinforce the value of the structured dose information in radiotherapy plan analytics and suggest that advanced language embeddings may offer limited additional benefits when dose metrics are already well-curated and standardized.

    \subsection{Interpretability and Modular System Design}
        Our results illustrate that a retrieval-augmented and modular approach yields both clinically accurate and highly interpretable treatment plan evaluations. By grounding percentile predictions in structured and numerical similarity rather than relying on opaque embedding-based reasoning, our system avoids the pitfalls of ``black-box" AI models, such as unpredictable behaviors and clinician mistrust, while still leveraging LLMs in a controlled, tool-mediated manner. This aligns with broader trends emphasizing the necessity of transparency and explainability in medical AI systems, especially when high-stakes decisions are involved \cite{frasca2024explainable, hildt2025role}. In contrast to many end-to-end learning systems, our design enables traceability: each prediction can be traced back to the specific reference plans and associated constraints. This model of “glass-box” decision support resonates with recommendations from the clinical AI literature, which emphasize auditability and explainable outputs to promote safety and clinician acceptance \cite{frasca2024explainable, kim2024xai}. The dominant role of numerical DVH-based similarity—versus semantic embeddings—in our retrieval configurations suggests that structured clinical features are often more informative for risk estimation than free-text or metadata encoding. This observation echoes findings from case-based reasoning frameworks in healthcare, where feature-based nearest neighbor retrieval has shown strong performance in diagnosis and treatment recommendation tasks \cite{campillo2013improving, kim2024xai}. Finally, our system’s separation of retrieval, constraint logic, and LLM-based reasoning naturally supports iterative refinements and human-in-the-loop workflows. This design not only enhances safety and adaptability but also better aligns with regulatory and institutional requirements governing AI in healthcare deployment \cite{abgrall2024should, okada2023explainable}.
        
    \subsection{Limitations and Methodological Considerations}
        While our system demonstrates high accuracy and robustness, several limitations must be acknowledged. Most notably, the dataset size remains constrained due to limited patient enrollment in clinical trials and the inherent challenges of accessing radiotherapy plans with appropriate patient consent. This restricts the diversity of case presentations and may limit generalizability to rare or highly individualized treatment contexts. In addition, although we tested a range of SentenceTransformer backbone models, all models exhibited minimal reliance on sentence embeddings, favoring numerical dose metrics instead. This suggests that current general-purpose LLMs may not fully capture radiotherapy-specific semantic nuances, and we did not explore more targeted alternatives such as domain-adapted encoders or LLMs fine-tuned based on medical corpora. This choice was driven by time and resource constraints. In future work, we plan to investigate the use of clinical-domain encoders (e.g., BioBERT, PubMedBERT) or fine-tuning strategies tailored to radiotherapy-specific domain knowledge. We hypothesize that such models may improve performance in settings where semantic structure names play a larger role or where DVH index information alone is insufficient for disambiguation. The scalarized loss function used to fine tune retrieval hyperparameters played a vital role in consolidating multiple evaluation metrics into a single, reproducible optimization target. However, because there is no formally defined similarity function for percentile estimation in this domain, the extent to which this loss captures clinical relevance or predictive confidence remains uncertain. To address this, we plan to benchmark the scalarized loss function against downstream clinical outcomes and conduct calibration studies to evaluate how well its predictions align with expert interpretation.

        While our normalized DVH metrics are derived using protocol-defined endpoints, we acknowledge that differences in calculation methods, such as dose grid resolution, interpolation, or ROI definitions, may lead to discrepancies compared to values observed in clinical systems. These differences could affect constraint evaluations for certain plans. Although we have aimed for consistent implementation across all records, further harmonization with clinical systems and validation against source planning data may help reduce such inconsistencies.
        
    \subsection{Future Work and Clinical Validation}
        Moving forward, we aim to extend validation beyond retrospective consistency checks by conducting prospective user studies involving radiation oncologists. These studies will assess whether system-generated plan evaluations align with clinical judgment and improve decision-making workflows. In particular, we see strong potential for our system to support adaptive treatment planning scenarios, where clinicians must select or revise plans in near real-time while the patient is on the treatment table. Since our retrieval-augmented framework can rapidly surface clinically similar reference plans and provide interpretable constraint-based summaries, it may offer decision support during tight time windows when adaptive planning is necessary. Future evaluations will explore whether the system's latency, interpretability, and accuracy are sufficient to assist in such time-sensitive clinical settings. Clinician feedback will also inform the development of more principled retrieval objectives and may motivate adjustments to feature weighting schemes or similarity metrics. In parallel, we will explore automated prompt engineering methods, such as reinforcement learning with human feedback (RLHF) or contrastive prompt optimization, to enhance robustness under diverse inputs. Finally, future versions of our system may incorporate richer clinical contexts, such as physician notes, anatomical imaging data, or planning intent, to support more holistic and personalized radiotherapy treatment plan evaluation.

\section{Conclusion}
    We presented a retrieval-augmented generation LLaMA 4 109B system for evaluating radiotherapy treatment plans in a protocol-aware and interpretable manner. By combining structured population-based scoring, numerical and semantic retrieval, constraint checking, and LLM–driven summarization, our system offers a transparent, modular framework for clinical radiotherapy plan assessment.

    Our experiments demonstrate that the system achieves high accuracy in both retrieval-based percentile prediction and constraint violation identification, with perfect agreement with computed values from individual modules in all evaluated cases. Notably, we found that normalized dose metrics play a dominant role in determining plan similarity with minimal contribution from sentence embeddings, an insight that underscores the value of structured clinical features over free-text representations in this domain.
    
\section*{Author Contribution}
    J.C. developed the methodology, implemented the codebase, conducted the experiments, and drafted the manuscript. P.W. provided guidance on large language model tool development and assisted with data acquisition. J.H. developed the tool for extracting dose-volume histogram metrics from raw DICOM files. L.S. contributed to manuscript revision and coordinated communications with clinical collaborators. M.L.H., B.A.P., S.A.V., T.T.S., W.W.W., N.Y.Y., S.E.S., J.R.N., S.R.K., J.-C.M.R., S.H.P., L.A.M., and C.A.V. contributed clinical expertise as treating physicians or protocol leads, reviewed the system for clinical relevance, and ensured adherence to trial guidelines. W.L. supervised the study, provided strategic guidance, and contributed to manuscript refinement. All authors reviewed and approved the final manuscript.
    
\section*{Declarations}

    \subsection*{Acknowledgments}

    None.
    
    \subsection*{Funding}

    This research was supported by NIH/BIBIB R01EB293388, by NIH/NCI R01CA280134, by the Eric \& Wendy Schmidt Fund for AI Research \& Innovation, and by the Kemper Marley Foundation.

\bibliographystyle{unsrt}
\bibliography{references}

\begin{thebibliography}{10}

\bibitem{gardner2019modern}
Stephen~J Gardner, Joshua Kim, and Indrin~J Chetty.
\newblock Modern radiation therapy planning and delivery.
\newblock {\em Hematology/Oncology Clinics}, 33(6):947--962, 2019.

\bibitem{deng2021critical}
Wei Deng, Yunze Yang, Chenbin Liu, Martin Bues, Radhe Mohan, William~W Wong, Robert~H Foote, Samir~H Patel, and Wei Liu.
\newblock A critical review of let-based intensity-modulated proton therapy plan evaluation and optimization for head and neck cancer management.
\newblock {\em International Journal of Particle Therapy}, 8(1):36--49, 2021.

\bibitem{zhang2011parameterization}
Xiaodong Zhang, Wei Liu, Yupeng Li, Xiaoqiang Li, Michelle Quan, Radhe Mohan, Aman Anand, Narayan Sahoo, Michael Gillin, and Xiaorong~R Zhu.
\newblock Parameterization of multiple bragg curves for scanning proton beams using simultaneous fitting of multiple curves.
\newblock {\em Physics in Medicine \& Biology}, 56(24):7725, 2011.

\bibitem{schild2014proton}
Steven~E Schild, William~G Rule, Jonathan~B Ashman, Sujay~A Vora, Sameer Keole, Aman Anand, Wei Liu, and Martin Bues.
\newblock Proton beam therapy for locally advanced lung cancer: A review.
\newblock {\em World journal of clinical oncology}, 5(4):568, 2014.

\bibitem{li2012dynamically}
Heng Li, Yupeng Li, Xiaodong Zhang, Xiaoqiang Li, Wei Liu, Michael~T Gillin, and X~Ronald Zhu.
\newblock Dynamically accumulated dose and 4d accumulated dose for moving tumors.
\newblock {\em Medical physics}, 39(12):7359--7367, 2012.

\bibitem{shan2020intensity}
Jie Shan, Yunze Yang, Steven~E Schild, Thomas~B Daniels, William~W Wong, Mirek Fatyga, Martin Bues, Terence~T Sio, and Wei Liu.
\newblock Intensity-modulated proton therapy (impt) interplay effect evaluation of asymmetric breathing with simultaneous uncertainty considerations in patients with non-small cell lung cancer.
\newblock {\em Medical physics}, 47(11):5428--5440, 2020.

\bibitem{hernandez2020plan}
Victor Hernandez, Christian~R{\o}nn Hansen, Lamberto Widesott, Anna B{\"a}ck, Richard Canters, Marco Fusella, Julia G{\"o}tstedt, Diego Jurado-Bruggeman, Nobutaka Mukumoto, Laura~Patricia Kaplan, et~al.
\newblock What is plan quality in radiotherapy? the importance of evaluating dose metrics, complexity, and robustness of treatment plans.
\newblock {\em Radiotherapy and Oncology}, 153:26--33, 2020.

\bibitem{geurts2022aapm}
Mark~W Geurts, Dustin~J Jacqmin, Lindsay~E Jones, Stephen~F Kry, Dimitris~N Mihailidis, Jared~D Ohrt, Timothy Ritter, Jennifer~B Smilowitz, and Nicholai~E Wingreen.
\newblock Aapm medical physics practice guideline 5. b: Commissioning and qa of treatment planning dose calculations—megavoltage photon and electron beams.
\newblock {\em Journal of applied clinical medical physics}, 23(9):e13641, 2022.

\bibitem{leone2024visualization}
Alexandra~O Leone, Abdallah~SR Mohamed, Clifton~D Fuller, Christine~B Peterson, Adam~S Garden, Anna Lee, Lauren~L Mayo, Amy~C Moreno, Jay~P Reddy, Karen Hoffman, et~al.
\newblock A visualization and radiation treatment plan quality scoring method for triage in a population-based context.
\newblock {\em Advances in Radiation Oncology}, 9(8):101533, 2024.

\bibitem{wang2025feasibility}
Qingxin Wang, Zhongqiu Wang, Minghua Li, Xinye Ni, Rong Tan, Wenwen Zhang, Maitudi Wubulaishan, Wei Wang, Zhiyong Yuan, Zhen Zhang, et~al.
\newblock A feasibility study of automating radiotherapy planning with large language model agents.
\newblock {\em Physics in Medicine \& Biology}, 70(7):075007, 2025.

\bibitem{hussein2018automation}
Mohammad Hussein, Ben~JM Heijmen, Dirk Verellen, and Andrew Nisbet.
\newblock Automation in intensity modulated radiotherapy treatment planning—a review of recent innovations.
\newblock {\em The British journal of radiology}, 91(1092):20180270, 2018.

\bibitem{cao2022knowledge}
Wenhua Cao, Mary Gronberg, Adenike Olanrewaju, Thomas Whitaker, Karen Hoffman, Carlos Cardenas, Adam Garden, Heath Skinner, Beth Beadle, and Laurence Court.
\newblock Knowledge-based planning for the radiation therapy treatment plan quality assurance for patients with head and neck cancer.
\newblock {\em Journal of applied clinical medical physics}, 23(6):e13614, 2022.

\bibitem{hansen2022plan}
Christian~R{\o}nn Hansen, Mohammad Hussein, Uffe Bernchou, Ruta Zukauskaite, and David Thwaites.
\newblock Plan quality in radiotherapy treatment planning--review of the factors and challenges.
\newblock {\em Journal of Medical Imaging and Radiation Oncology}, 66(2):267--278, 2022.

\bibitem{patel2021radiotherapy}
Ganeshkumar Patel, Abhijit Mandal, Ravindra Shende, and Avinav Bharati.
\newblock Radiotherapy plan evaluation indices: A dosimetrical suitability check.
\newblock {\em Journal of Cancer Research and Therapeutics}, 17(2):455--462, 2021.

\bibitem{liu2017exploratory}
Wei Liu, Samir~H Patel, Daniel~P Harrington, Yanle Hu, Xiaoning Ding, Jiajian Shen, Michele~Y Halyard, Steven~E Schild, William~W Wong, Gary~E Ezzell, et~al.
\newblock Exploratory study of the association of volumetric modulated arc therapy (vmat) plan robustness with local failure in head and neck cancer.
\newblock {\em Journal of applied clinical medical physics}, 18(4):76--83, 2017.

\bibitem{liu2016robustness}
Wei Liu, Samir~H Patel, Jiajian~Jason Shen, Yanle Hu, Daniel~P Harrington, Xiaoning Ding, Michele~Y Halyard, Steven~E Schild, William~W Wong, Gary~A Ezzell, et~al.
\newblock Robustness quantification methods comparison in volumetric modulated arc therapy to treat head and neck cancer.
\newblock {\em Practical radiation oncology}, 6(6):e269--e275, 2016.

\bibitem{kang2019impact}
Yixiu Kang, Jiajian Shen, Wei Liu, Paige~A Taylor, Hunter~S Mehrens, Xiaoning Ding, Yanle Hu, Erik Tryggestad, Sameer~R Keole, Steven~E Schild, et~al.
\newblock Impact of planned dose reporting methods on gamma pass rates for iroc lung and liver motion phantoms treated with pencil beam scanning protons.
\newblock {\em Radiation Oncology}, 14(1):108, 2019.

\bibitem{liu2023artificial}
Chenbin Liu, Zhengliang Liu, Jason Holmes, Lu~Zhang, Lian Zhang, Yuzhen Ding, Peng Shu, Zihao Wu, Haixing Dai, Yiwei Li, et~al.
\newblock Artificial general intelligence for radiation oncology.
\newblock {\em Meta-radiology}, 1(3):100045, 2023.

\bibitem{wahl2020analytical}
Niklas Wahl, Philipp Hennig, Hans-Peter Wieser, and Mark Bangert.
\newblock Analytical probabilistic modeling of dose-volume histograms.
\newblock {\em Medical physics}, 47(10):5260--5273, 2020.

\bibitem{mayo2017incorporating}
Charles~S Mayo, John Yao, Avraham Eisbruch, James~M Balter, Dale~W Litzenberg, Martha~M Matuszak, Marc~L Kessler, Grant Weyburn, Carlos~J Anderson, Dawn Owen, et~al.
\newblock Incorporating big data into treatment plan evaluation: Development of statistical dvh metrics and visualization dashboards.
\newblock {\em Advances in radiation oncology}, 2(3):503--514, 2017.

\bibitem{engberg2017explicit}
Lovisa Engberg, Anders Forsgren, Kjell Eriksson, and Bj{\"o}rn H{\aa}rdemark.
\newblock Explicit optimization of plan quality measures in intensity-modulated radiation therapy treatment planning.
\newblock {\em Medical Physics}, 44(6):2045--2053, 2017.

\bibitem{zhang2020direct}
Tianfang Zhang, Rasmus Bokrantz, and Jimmy Olsson.
\newblock Direct optimization of dose--volume histogram metrics in radiation therapy treatment planning.
\newblock {\em Biomedical Physics \& Engineering Express}, 6(6):065018, 2020.

\bibitem{pyakuryal2010computational}
Anil Pyakuryal, W~Kenji Myint, Mahesh Gopalakrishnan, Sunyoung Jang, Jerilyn~A Logemann, and Bharat~B Mittal.
\newblock A computational tool for the efficient analysis of dose-volume histograms for radiation therapy treatment plans.
\newblock {\em Journal of Applied Clinical Medical Physics}, 11(1):137--157, 2010.

\bibitem{mesinovic2025explainability}
Munib Mesinovic, Peter Watkinson, and Tingting Zhu.
\newblock Explainability in the age of large language models for healthcare.
\newblock {\em Communications Engineering}, 4(1):128, 2025.

\bibitem{hu2025systematic}
Di~Hu, Yawen Guo, Yiliang Zhou, Lidia Flores, and Kai Zheng.
\newblock A systematic review of early evidence on generative ai for drafting responses to patient messages.
\newblock {\em npj Health Systems}, 2(1):27, 2025.

\bibitem{wang2023mint}
Xingyao Wang, Zihan Wang, Jiateng Liu, Yangyi Chen, Lifan Yuan, Hao Peng, and Heng Ji.
\newblock Mint: Evaluating llms in multi-turn interaction with tools and language feedback.
\newblock {\em arXiv preprint arXiv:2309.10691}, 2023.

\bibitem{RN1787}
H.~Dai, Z.~Liu, W.~Liao, X.~Huang, Y.~Cao, Z.~Wu, L.~Zhao, S.~Xu, F.~Zeng, W.~Liu, N.~Liu, S.~Li, D.~Zhu, H.~Cai, L.~Sun, Q.~Li, D.~Shen, T.~Liu, and X.~Li.
\newblock Auggpt: Leveraging chatgpt for text data augmentation.
\newblock {\em IEEE Transactions on Big Data}, 11(03):907--918, 2025.

\bibitem{RN1782}
Yuexing Hao, Jason Holmes, Jared Hobson, Alexandra Bennett, Elizabeth~L. McKone, Daniel~K. Ebner, David~M. Routman, Satomi Shiraishi, Samir~H. Patel, Nathan~Y. Yu, Chris~L. Hallemeier, Brooke~E. Ball, Mark Waddle, and Wei Liu.
\newblock Retrospective comparative analysis of prostate cancer in-basket messages: Responses from closed-domain large language models versus clinical teams.
\newblock {\em Mayo Clinic Proceedings: Digital Health}, 3(1):100198, 2025.

\bibitem{RN1790}
Yuexing Hao, Zhiwen Qiu, Jason Holmes, Corinna~E. Löckenhoff, Wei Liu, Marzyeh Ghassemi, and Saleh Kalantari.
\newblock Large language model integrations in cancer decision-making: a systematic review and meta-analysis.
\newblock {\em npj Digital Medicine}, 8(1):450, 2025.

\bibitem{RN1753}
J.~Holmes, L.~Zhang, Y.~Ding, H.~Feng, Z.~Liu, T.~Liu, W.~W. Wong, S.~A. Vora, J.~B. Ashman, and W.~Liu.
\newblock Benchmarking a foundation large language model on its ability to relabel structure names in accordance with the american association of physicists in medicine task group-263 report.
\newblock {\em Pract Radiat Oncol}, 2024.

\bibitem{RN1169}
Z.~Liu, M.~He, Z.~Jiang, Z.~Wu, H.~Dai, L.~Zhang, S.~Luo, T.~Han, X.~Li, X.~Jiang, D.~Zhu, X.~Cai, B.~Ge, W.~Liu, J.~Liu, D.~Shen, and T.~Liu.
\newblock Survey on natural language processing in medical image analysis.
\newblock {\em Zhong Nan Da Xue Xue Bao Yi Xue Ban}, 47(8):981--993, 2022.

\bibitem{RN1789}
Zhengliang Liu, Yiwei Li, Peng Shu, Aoxiao Zhong, Hanqi Jiang, Yi~Pan, Longtao Yang, Chao Ju, Zihao Wu, Chong Ma, Cheng Chen, Sekeun Kim, Haixing Dai, Lin Zhao, Lichao Sun, Dajiang Zhu, Jun Liu, Wei Liu, Dinggang Shen, Quanzheng Li, Tianming Liu, and Xiang Li.
\newblock Radiology-gpt: A large language model for radiology.
\newblock {\em Meta-Radiology}, 3(2):100153, 2025.

\bibitem{RN1559}
Z.~Liu, L.~Zhang, Z.~Wu, X.~Yu, C.~Cao, H.~Dai, N.~Liu, J.~Liu, W.~Liu, Q.~Li, D.~Shen, X.~Li, D.~Zhu, and T.~Liu.
\newblock Surviving chatgpt in healthcare.
\newblock {\em Front Radiol}, 3:1224682, 2023.

\bibitem{RN1553}
Zhengliang Liu, Aoxiao Zhong, Yiwei Li, Longtao Yang, Chao Ju, Zihao Wu, Chong Ma, Peng Shu, Cheng Chen, Sekeun Kim, Haixing Dai, Lin Zhao, Dajiang Zhu, Jun Liu, Wei Liu, Dinggang Shen, Quanzheng Li, Tianming Liu, and Xiang Li.
\newblock Tailoring large language models to radiology: A preliminary approach to llm adaptation for a highly specialized domain.
\newblock In Xiaohuan Cao, Xuanang Xu, Islem Rekik, Zhiming Cui, and Xi~Ouyang, editors, {\em Machine Learning in Medical Imaging}, pages 464--473. Springer Nature Switzerland, 2024.

\bibitem{RN1389}
Saed Rezayi, Haixing Dai, Zhengliang Liu, Zihao Wu, Akarsh Hebbar, Andrew~H. Burns, Lin Zhao, Dajiang Zhu, Quanzheng Li, Wei Liu, Sheng Li, Tianming Liu, and Xiang Li.
\newblock Clinicalradiobert: Knowledge-infused few shot learning for clinical notes named entity recognition.
\newblock In Chunfeng Lian, Xiaohuan Cao, Islem Rekik, Xuanang Xu, and Zhiming Cui, editors, {\em Machine Learning in Medical Imaging}, pages 269--278. Springer Nature Switzerland, 2022.

\bibitem{RN1785}
Peilong Wang, Jason Holmes, Zhengliang Liu, Dequan Chen, Tianming Liu, Jiajian Shen, and Wei Liu.
\newblock A recent evaluation on the performance of llms on radiation oncology physics using questions of randomly shuffled options.
\newblock {\em Frontiers in Oncology}, Volume 15 - 2025, 2025.

\bibitem{budnikov2025generalization}
Mikhail Budnikov, Anna Bykova, and Ivan~P Yamshchikov.
\newblock Generalization potential of large language models.
\newblock {\em Neural Computing and Applications}, 37(4):1973--1997, 2025.

\bibitem{goodell2025large}
Alex~J Goodell, Simon~N Chu, Dara Rouholiman, and Larry~F Chu.
\newblock Large language model agents can use tools to perform clinical calculations.
\newblock {\em npj Digital Medicine}, 8(1):163, 2025.

\bibitem{nusrat2025autonomous}
Humza Nusrat, Bing Luo, Ryan Hall, Joshua Kim, Hassan Bagher-Ebadian, Anthony Doemer, Benjamin Movsas, and Kundan Thind.
\newblock Autonomous radiotherapy treatment planning using dola: A privacy-preserving, llm-based optimization agent.
\newblock {\em arXiv preprint arXiv:2503.17553}, 2025.

\bibitem{amugongo2025retrieval}
Lameck~Mbangula Amugongo, Pietro Mascheroni, Steven Brooks, Stefan Doering, and Jan Seidel.
\newblock Retrieval augmented generation for large language models in healthcare: A systematic review.
\newblock {\em PLOS Digital Health}, 4(6):e0000877, 2025.

\bibitem{wada2025retrieval}
Akihiko Wada, Yuya Tanaka, Mitsuo Nishizawa, Akira Yamamoto, Toshiaki Akashi, Akifumi Hagiwara, Yayoi Hayakawa, Junko Kikuta, Keigo Shimoji, Katsuhiro Sano, et~al.
\newblock Retrieval-augmented generation elevates local llm quality in radiology contrast media consultation.
\newblock {\em npj Digital Medicine}, 8(1):395, 2025.

\bibitem{liu2024automated}
Sheng Liu, Oscar Pastor-Serrano, Yizheng Chen, Matthew Gopaulchan, Weixin Liang, Mark Buyyounouski, Erqi Pollom, Quynh-Thu Le, Michael~Francis Gensheimer, Peng Dong, et~al.
\newblock Automated radiotherapy treatment planning guided by gpt-4vision.
\newblock {\em Physics in Medicine and Biology}, 2024.

\bibitem{xu2024crp}
Kehan Xu, Kun Zhang, Jingyuan Li, Wei Huang, and Yuanzhuo Wang.
\newblock Crp-rag: A retrieval-augmented generation framework for supporting complex logical reasoning and knowledge planning.
\newblock {\em Electronics}, 14(1):47, 2024.

\bibitem{liu2023radonc}
Zhengliang Liu, Peilong Wang, Yiwei Li, Jason Holmes, Peng Shu, Lian Zhang, Chenbin Liu, Ninghao Liu, Dajiang Zhu, Xiang Li, et~al.
\newblock Radonc-gpt: A large language model for radiation oncology.
\newblock {\em arXiv preprint arXiv:2309.10160}, 2023.

\bibitem{frasca2024explainable}
Maria Frasca, Davide La~Torre, Gabriella Pravettoni, and Ilaria Cutica.
\newblock Explainable and interpretable artificial intelligence in medicine: a systematic bibliometric review.
\newblock {\em Discover Artificial Intelligence}, 4(1):15, 2024.

\bibitem{hildt2025role}
Elisabeth Hildt.
\newblock What is the role of explainability in medical artificial intelligence? a case-based approach.
\newblock {\em Bioengineering}, 12(4):375, 2025.

\bibitem{kim2024xai}
Se~Young Kim, Dae~Ho Kim, Min~Ji Kim, Hyo~Jin Ko, and Ok~Ran Jeong.
\newblock Xai-based clinical decision support systems: a systematic review.
\newblock {\em Applied Sciences}, 14(15):6638, 2024.

\bibitem{campillo2013improving}
Boris Campillo-Gimenez, Wassim Jouini, Sahar Bayat, and Marc Cuggia.
\newblock Improving case-based reasoning systems by combining k-nearest neighbour algorithm with logistic regression in the prediction of patients’ registration on the renal transplant waiting list.
\newblock {\em PLoS One}, 8(9):e71991, 2013.

\bibitem{abgrall2024should}
Gw{\'e}nol{\'e} Abgrall, Andre~L Holder, Zaineb Chelly~Dagdia, Karine Zeitouni, and Xavier Monnet.
\newblock Should ai models be explainable to clinicians?
\newblock {\em Critical Care}, 28(1):301, 2024.

\bibitem{okada2023explainable}
Yohei Okada, Yilin Ning, and Marcus Eng~Hock Ong.
\newblock Explainable artificial intelligence in emergency medicine: an overview.
\newblock {\em Clinical and experimental emergency medicine}, 10(4):354, 2023.

\end{thebibliography}

\end{document}